\documentclass[letterpaper]{article}
\usepackage{aaai}
\usepackage{times}
\usepackage{helvet}
\usepackage{courier}
\usepackage{enumitem}
\setlist{nosep}

\usepackage[dvipsnames]{xcolor}
\usepackage{verbatim}
\usepackage{url}
\usepackage{bm}
\usepackage{booktabs}
\usepackage{graphicx}
\usepackage{multirow}
\usepackage{amssymb}
\usepackage{amsmath}
\usepackage{soul}
\newcommand{\ct}[1]{\citeauthor{#1} \shortcite{#1}}

\begin{document}
%
\title{A task in a suit and a tie: paraphrase generation with semantic augmentation\thanks{Work done while the first author was an intern at Google AI.}}
\author{
Su Wang$^{1}$ \ \ \ \ \ \  Rahul Gupta$^{2}$\ \ \ \ \ \  Nancy Chang$^{2}$ \ \ \ \ \ \ Jason Baldridge$^{2}$\\\\
The University of Texas at Austin$^{1}$\\
Google AI Language$^{2}$\\
\texttt{shrekwang@utexas.edu; \{grahul, ncchang, jasonbaldridge\}@google.com}
}

\maketitle
\begin{abstract}
\begin{quote}


Paraphrasing is rooted in semantics. We show the effectiveness of transformers \cite{Vaswani:17} for paraphrase generation and further improvements by incorporating PropBank labels via a multi-encoder. Evaluating on MSCOCO and WikiAnswers, we find that transformers are fast and effective, and that semantic augmentation for both transformers and LSTMs leads to sizable 2-3 point gains in \texttt{BLEU}, \texttt{METEOR} and \texttt{TER}. More importantly, we find surprisingly large gains on human evaluations compared to previous models. Nevertheless, manual inspection of generated paraphrases reveals ample room for improvement: even our best model produces human-acceptable paraphrases for only 28\% of captions from the CHIA dataset \cite{Sharma:18}, and it fails spectacularly on sentences from Wikipedia. Overall, these results point to the potential for incorporating semantics in the task while highlighting the need for stronger evaluation.

\end{quote}
\end{abstract}

\section{Introduction}
\label{sec:intro}

Paraphrasing is at its core a semantic task: restate one phrase $s$ as another $s'$ with approximately the same meaning. High-precision, domain-general paraphrase generation can benefit many natural language processing tasks \cite{Madnani:10b}. For example, paraphrases can help diversify responses of dialogue assistants \cite{Shah:18}, augment machine translation training data \cite{Fader:14} and extend coverage of semantic parsers \cite{berant-liang:2014:P14-1}. But what makes a good paraphrase? Whether two phrases have ``the same'' meaning leaves much room for interpretation. As Table \ref{tab:variation} illustrates, paraphrases can involve lexical and syntactic variation with differing degrees of semantic fidelity. Should a paraphrase yield precisely the same inferences as the original? Should it refer to the same entities, in similar ways, with the same detail? How much lexical overlap between paraphrases is desirable?

\begin{table}[!t]
\begin{center}
\scalebox{0.85}{
\begin{tabular}{cl}
Source & Sue sold her car to Bob. \\
\midrule
& Sue sold her auto to Bob. \\
A & Sue sold Bob her car. \\
& Bob was sold a car by Sue. \\
\midrule
& Bob bought a vehicle from Sue. \\
B& Bob bought Sue's car (from her). \\
& Bob paid Sue for her automobile. \\
& Sue let Bob buy her auto. \\
\midrule
C& Sue gave Bob a good deal on her Honda. \\
& Bob got a bargain on Sue's Honda. \\
\newline
\end{tabular}}
\caption{{\small Paraphrases of a source sentence with increasing syntactic and semantic variability: (A) mild lexical and syntactic variation; (B) moderate lexical and syntactic variation plus perspective shifting; (C) even more lexical and syntactic variation, with greater semantic license.}}\label{tab:variation}
\end{center}
\end{table}



Much prior work on paraphrasing addresses these questions directly by exploiting linguistic knowledge, including handcrafted rules \cite{McKeown:83}, shallow linguistic features \cite{Zhao:09} and syntactic and semantic information \cite{Kozlowski:03,Ellsworth:07}. These systems, while difficult to scale and typically restricted in domain, nonetheless produce paraphrases that are qualitatively close to what humans expect. For example, \ct{Ellsworth:07}'s rule-based system uses frame semantic information \cite{fillmore82:_frame} to paraphrase \emph{I want your opinion} as \emph{Your opinion is desired}, where the two sentences evoke the \textsc{Desiring} frame from different perspectives.

Many recent paraphrase generation systems sidestep explicit questions of meaning, focusing instead on implicitly learning representations by training deep neural networks (see next section) on datasets containing paraphrase pairs. This is essentially the same approach as modern sequence-to-sequence machine translation (MT), another intrinsically semantic task for which deep learning on large-scale end-to-end data has yielded large gains in quality.
Unlike machine translation, however, the training data currently available is much smaller and more domain-specific for paraphrasing. Despite incremental improvements on automatically scored metrics, the paraphrasing quality of state-of-the-art systems fall far short of what's needed for most applications.

This work investigates a combination of these two approaches. Can adding structured linguistic information to large-scale deep learning systems improve the quality and generalizability of paraphrase generation while remaining scalable and efficient?\footnote{While our models are resource-dependent, we believe it is reasonable to utilize existing resources to introduce semantic structure. As we will show, this approach holds great promise when combined with a general-purpose semantic parser.} We present a fast-converging and data-efficient paraphrase generator that exploits structured semantic knowledge within the Transformer framework \cite{Vaswani:17,Xiong:18}. Our system outperforms state-of-the-art systems \cite{Prakash:16,Gupta:18} on automatic evaluation metrics and data benchmarks, with improvements of 2-3 \texttt{BLEU}, \texttt{METEOR} and \texttt{TER} points on both MSCOCO and WikiAnswers. 

Despite these substantial gains, manual examination of our system's output on out-of-domain sentences reveals serious problems. Models with high \texttt{BLEU} scores often paraphrased unfamiliar inputs as {\it a man in a suit and a tie...} (see Table \ref{tab:suit-tie}). This striking pattern suggests we need more extensive evaluation to better characterize model behavior. We thus obtain human judgments on the outputs of several models to measure both overall quality and relative differences between the models. Interestingly, we find that some models with similar \texttt{BLEU} scores differ widely in human evaluations: e.g. an LSTM-based model with 41.1 \texttt{BLEU} score gets 36.4\% acceptability, while a Transformer model with 41.0 \texttt{BLEU} score rates 45.6\%. These and our other human evaluations and observations imply that automatic metrics are useful for model development for paraphrase generation research, but not sufficient for final comparison. 

Our results also indicate considerable headroom for granting semantics a greater role throughout---in representation, architecture design and task evaluation. This emphasis is in line with other recent and related work, including entailment in paraphrasing \cite{pavlick:etal:2015}, paraphrase identification with Abstract Meaning Representations \cite{issa:etal:2018} and using semantic role labeling in machine translation \cite{marcheggiani:etal:2018}. It should be noted that this work is only a first step toward paraphrasing at the quality exemplified in Table \ref{tab:variation}. We believe that achieving that level requires much more fundamental work to define appropriate tasks, metrics and optimization objectives.

\section{Data and Evaluation}
\label{sec:data-and-eval}

\begin{table}[!t]
\begin{center}
\scalebox{0.85}{
\begin{tabular}{lccc}
Dataset & Train & Test & Vocab \\
\midrule
\texttt{MSCOCO} & 331,163 & 162,023 & 30,332 \\
\texttt{WikiAnswers} & 1,826,492 & 500,000 & 86,334\\
\end{tabular}}
\caption{{\small Data statistics}}\label{tab:data-stats}
\end{center}
\end{table}

\textbf{Data}. We use MSCOCO \cite{Lin:14} and WikiAnswers \cite{Fader:13} as our main datasets for training and evaluation, as prepared by \ct{Prakash:16}. MSCOCO contains 500K+ paraphrases pairs created by crowdsourcing captions for a set of $\sim$120K images (e.g., a horse grazing in a meadow). Each image is captioned by 5 people with a (single/multi-clause) sentence. WikiAnswers has 2.3M+ questions pairs marked by users of the WikiAnswers website\footnote{\url{wiki.answers.com}} as similar or duplicate (e.g. \textit{where is Mexico City?} and \textit{in which country is Mexico City located?}). See Table \ref{tab:data-stats} for basic statistics of both datasets.



\medskip
\noindent
\textbf{Evaluation}. Paraphrasing systems have been evaluated using purely human methods \cite{Barzilay:01,Bannard:05}, human-in-the-loop semi-automatic methods \cite[ParaMetric]{Callison-Burch:08}, resource-dependent automatic methods \cite[PEM]{Liu:10}, and cost-efficient human evaluation aided by automatic methods \cite{Chaganty:18}. 

Recent neural paraphrasing systems \cite{Prakash:16,Gupta:18} adopt automatic evaluation measures commonly used in MT, citing good correlation with human judgment \cite{Madnani:10a,Wubben:10}:
\texttt{BLEU} \cite{Papineni:02} encourages exact match between source and prediction by n-gram overlap; \texttt{METEOR} \cite{Lavie:07} also uses WordNet stems and synonyms; and \texttt{TER} \cite{Snover:06} includes the number of edits between source and prediction.

Such automatic metrics enable fast development cycles, but they are not sufficient for final quality assessment. For example, \ct{Chen:11} point out that MT metrics reward homogeneous predictions to the training target, which conflicts with a qualitative goal of good human-level paraphrasing: variation in wording. \ct{Chaganty:18} show that predictions receiving low scores from these metrics are not necessarily poor quality according to human evaluation. We thus complement these automated metrics with crowdsourced human evaluation.

\begin{figure}[!t]
\begin{center}
\includegraphics[width=80mm]{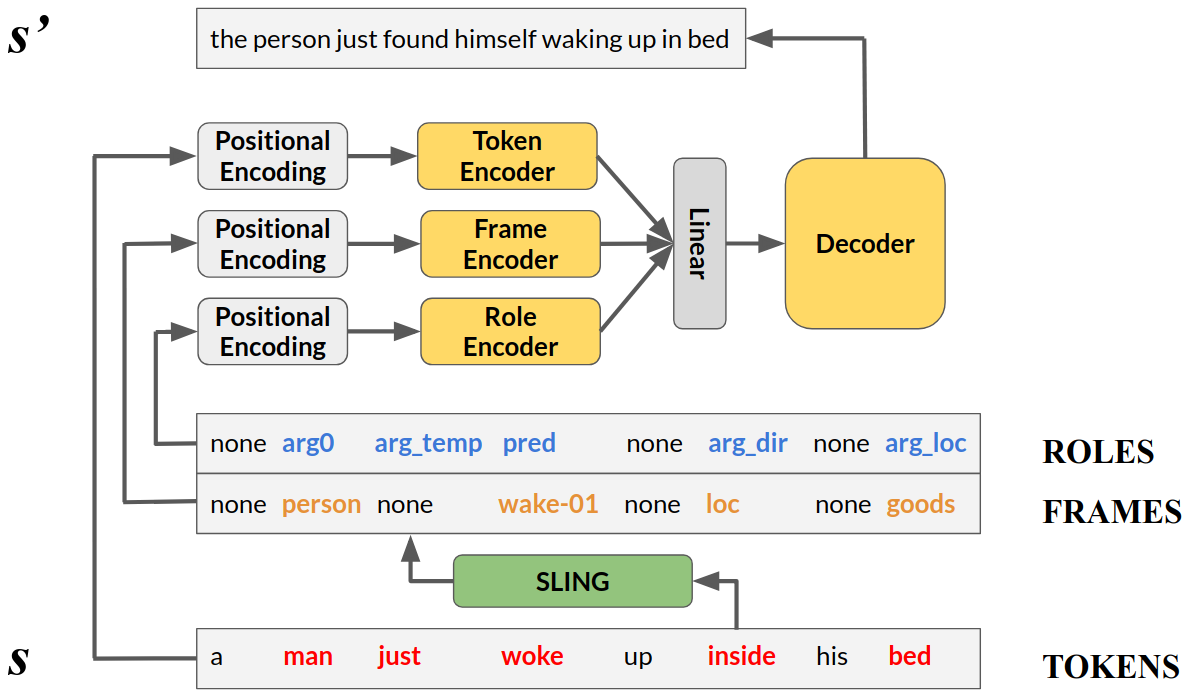}
\end{center}
\caption{{\small Multi-encoder Transformer (\textsc{transformer-pb}, \textsc{pb} denotes PropBank-style semantic annotations) with \texttt{SLING}. Each token is annotated with a frame and role label, resulting in three input channels. Each channel is fed into a separate \textsc{transformer} encoder, and results from the three encoders are merged with a linear layer for the decoder.}}\label{fig:model}
\end{figure}

\section{Models}
\label{sec:models}

Our core contribution is showing a straightforward way to inject semantic frames and roles into the encoders of seq2seq models. We also show that Transformers in particular are fast and effective for paraphrase generation.

\noindent
\textbf{Transformer}. The \textsc{transformer} attention-only seq2seq model proposed in \ct{Vaswani:17} has been shown to be a data-efficient, fast-converging architecture. It circumvents token-by-token encoding with a parallelized encoding step that uses token position information. Experimentally its sophisticated self-attention and target-to-source attention mechanism work robustly for long sequences (see the experimentation section). The basic building block of the \textsc{transformer} encoder is a multi-head attention layer followed by a feedforward layer, where both have residual links and layer norm \cite{Ba:16}:
{\small
\begin{align}
\bm{h} &= \texttt{LayerNorm}(\texttt{FFNN}(\bm{m})) + \bm{m} \\
\bm{m} &= \texttt{LayerNorm}(\texttt{MultiAttn}(\bm{x})) + \bm{x} \\
\bm{x} &= \texttt{PositionalEncoding}(s)
\end{align}}
where $\bm{h}$ denotes the encoding output of the block, $\bm{m}$ the encoding by the multi-attention layer, and $\bm{x}$ the positionally encoded word embeddings of the input sequence. $N$ encoding blocks are cloned to produce the final encoding outputs $\bm{h}_N$, where $\bm{h}_n = g_{n-1}(\bm{h}_{n-1})$, where $g_{n-1}$ is the previous encoding block. The decoding block is almost identical to the encoding block, with the addition of one more multi-attention layer before the feedforward layer, where the decoder attends to the encoding outputs $\bm{h}_N$. The decoder also has a number of clones of the decoding blocks, which is not necessarily equal to $N$. The final output of the decoder is projected through a linear layer followed by a softmax. 

\medskip
\noindent
\textbf{Encoding semantics}. We predict structured semantic representations with an off-the-shelf general-purpose frame-semantic parser \texttt{SLING} \cite{Ringgaard:17}. \texttt{SLING} is a neural transition-based semantic graph generator that incrementally generates a frame graph representing the meaning of its input text. \texttt{SLING}'s frame graph representation can homogeneously encode common semantic annotations such as entity annotations, semantic role labels (SRL), and more general frame and inter-frame annotations; the released \texttt{SLING} model\footnote{https://github.com/google/sling} provides entity, measurement, and PropBank-style SRL annotations out of the box. For example, for the sentence \emph{a man just woke up inside his bed}, the spans \emph{man} and \emph{woke} evoke frames of type \emph{person} and \emph{/propbank/wake-01}, respectively, and the latter frame links to the former via the role {\it arg0}, denoting that the \emph{person} frame is the subject of the predicate. Other spans in the sentence similarly evoke their corresponding frames and link with the appropriate roles (see Figure \ref{fig:model}). The end output of \texttt{SLING} is an inter-linked graph of frames.

Given an input sentence $s$, \texttt{SLING} first embeds and encodes the tokens of $s$ through a BiLSTM using lexical features. A feedforward Transition-Based Recurrent Unit (TBRU) then processes the token vectors one at a time, left to right. At each step, it shifts to the next token or makes an edit operation (or \emph{transition}) to the frame graph under construction (initially empty). Examples of such transitions are (a) evoking a new frame from a span starting at the current token, (b) linking two existing frames with a role, and (c) re-evoking an old frame for a new span (e.g.~as in coreference resolution). A full set of such operations is listed in \ct{Ringgaard:17}. To decide which operation to perform, \texttt{SLING} maintains and exploits a bounded priority queue of frames predicted so far, ordered by those most recently used/evoked. Since frame-graph construction boils down to a sequence of fairly general transitions, the model and architecture of \texttt{SLING} are independent of the annotation schema. The same model architecture can thus be trained and applied flexibly to diverse tasks, such as FrameNet parsing~\cite{Schneider:18}
or coreference resolution.

\texttt{SLING} offers two major benefits for our paraphrase generation task. First, since everything is cast in terms of frames and roles, multiple heterogenously annotations can be homogenously obtained in one model using \texttt{SLING}'s frame API. We use to access entity, measurements, and SRL frames output from a pre-trained \texttt{SLING} model. Second, the frame graph representation is powerful enough to capture many semantic tasks, so we can eventually try other ways of capturing semantics by changing the annotation schema, e.g. QA-SRL \cite{He:2015}, FrameNet annotations, open-domain facts, or coreference clusters.

Here, we incorporate \texttt{SLING}'s entity, measurement, and PropBank-SRL frames using a multi-encoder method. Taking \textsc{transformer} for example (Figure 1), we first use \texttt{SLING} to annotate the input sentence $s$ with frame and role labels, and transfer these labels to tokens, resulting in three aligned vectors for $s$: tokens, frames and roles. The three channels each have a separate \textsc{transformer} encoder. The encoders produce three sets of outputs which are merged with a linear layer before decoding. The multi-encoder \textsc{transformer} with the PropBank-style \texttt{SLING} annotations will be listed as \textsc{transformer-pb}.  

\medskip
\noindent
\textbf{Benchmarks}. We consider two state-of-the-art neural paraphrase generators: (1) a stacked residual LSTM by \ct{Prakash:16} (listed as \textsc{sr-lstm}), and (2) a nested variational LSTM by \ct{Gupta:18} (listed as \textsc{nv-lstm}). 

Let $s = \{\bm{w}_1,\dots,\bm{w}_K\}$ be an input sentence, where $\bm{w}_k$ denotes a word embedding. The \textsc{sr-lstm} encodes, with a bidirectional LSTM (Bi-LSTM), the sentence as a context vector $\bm{c}$ that encapsulates its representation. At each time-step at layer $l$, the LSTM cell takes input from the previous state, as well as a residual linking \cite{He:15} from the previous layer $l-1$:
{\small
\begin{align}
\bm{c} &= \texttt{Bi-LSTM}(s) \\
\bm{h}_t^l &= \texttt{LSTM-Cell}^l(\bm{h}_t^{l-1},\bm{h}_{t-1}^l) + \bm{w}_t 
\end{align}}
where $\bm{h}_t^{l}$ is the hidden state at layer $l$ and time-step $t$. \ct{Prakash:16} find the best balance between model capacity and generalization with 2-layered stacking. The decoder is initialized with the context vector $\bm{c}$ and has the same architecture as the encoder. Finally, the decoder maintains a standard attention \cite{Bahdanau:15} over the output states of the encoder.

\begin{table*}
\begin{center}
\scalebox{0.85}{
\begin{tabular}{lccccccc}
\multirow{2}{*}{Model} & \multicolumn{3}{c}{\texttt{MSCOCO}} & \multicolumn{3}{c}{\texttt{Wikianswers}} & \#Tokens/sec  \\
\cmidrule{2-7}
& \texttt{BLEU}$\uparrow$ & \texttt{METEOR}$\uparrow$ & \texttt{TER}$\downarrow$ & \texttt{BLEU}$\uparrow$ & \texttt{METEOR}$\uparrow$ & \texttt{TER}$\downarrow$ & (GPU) \\ 
\midrule
\textsc{sr-lstm} \cite{Prakash:16} & 36.7 & 27.3 & 52.3 & 37.0 & 32.2 & 27.0 & -\\
\textsc{nv-lstm} \cite{Gupta:18}  & 41.7 & 31.0 & 40.8 & - & - & - & -\\
\hline
\textsc{sr-lstm} (ours)$\dagger$ & 36.5 & 26.8 & 51.4 & 36.9 & 34.7 & 27.0 & 458 \\
\textsc{nv-lstm} (ours)$\dagger$ & 41.1 & 31.2 & 41.1 & 39.2 & 36.1 & 22.9 & 417 \\
\textsc{transformer} \cite{Vaswani:17} & 41.0 & 32.8 & 40.5 & 41.9 & 35.8 & 22.5 & \textbf{2,875} \\
\textsc{sr-lstm-pb} (ours) & 40.8 & 32.3 & 47.0 & 42.1 & 37.9 & 21.2 & 173 \\
\textsc{transformer-pb} (ours) & \textbf{44.0} & \textbf{34.7} & \textbf{37.1} & \textbf{43.9} & \textbf{38.7} & \textbf{19.4} & 2,650 \\
\end{tabular}}
\caption{{\small Results on \texttt{MSCOCO} and \textsc{Wikianswers} with length-15 sentence truncation. The arrows $\uparrow$ and $\downarrow$ indicate how the scores are interpreted: for \texttt{BLEU} and \texttt{METEOR}, the higher the better, for \texttt{TER}, the lower the better. The best result in a column is in boldface. The \#tokens/sec statistics: batch size 32.($\dagger$: We emphasize that these results are based on our reimplementations of the models described in the cited papers.)}}\label{tab:auto-results}
\end{center}
\end{table*}

For \textsc{nv-lstm}, the encoder consists of a sequence of two nested LSTMs: Let $s$ and $s'$ be a paraphrase input entry (source and target). The second nested LSTM takes $s'$ and the encoding of $s$ by the first nested LSTM. This results in the context vector $\bm{c}$:
{\small
\begin{equation}
\bm{c} = \texttt{LSTM}_2^{enc}(s', \texttt{LSTM}_1^{enc}(s))
\end{equation}}
The context vector is then fed through a standard variational reparameterization layer \cite{Kingma:14,Bowman:15} to produce encoding $\bm{z}$:
{\small
\begin{align}
\bm{z} &\sim \mathcal{N}(\bm{\mu},\bm{\sigma}) \\
\bm{\mu} &= \texttt{Linear}_{\mu}(\bm{h}) \\
\bm{\sigma} &= \texttt{Linear}_{\sigma}(\bm{h})
\end{align}}
The decoder also comprises two nested LSTMs: the first again encodes $s$ and produces its final state for the second LSTM for final decoding, which also conditions on $\bm{z}$:
{\small
\begin{equation}
\hat{s}' = \texttt{LSTM}_2^{dec}(\texttt{LSTM}_1^{dec}(s)\mid\bm{z})
\end{equation}}
where $\hat{s}'$ is the predicted sequence. 

To separate the contributions from architecture and semantics, we also implement a similar multi-encoder for the baseline \textsc{sr-lstm} (listed as \textsc{sr-lstm-pb}), where we again work with three copies of encoder Bi-LSTMs:
{\small
\begin{align}
\bm{c} &= \texttt{Linear}(\texttt{Concat}(\bm{c}_s,\bm{c}_f,\bm{c}_r)) \\
\bm{c}_s &= \texttt{Bi-LSTM}(s) \\
\bm{c}_f &= \texttt{Bi-LSTM}(f) \\
\bm{c}_r &= \texttt{Bi-LSTM}(r) \\
f, r &= \texttt{SLING}(s)
\end{align}}
where $f$ and $r$ are the same frame and role annotations from \texttt{SLING} as used in \textsc{transformer-pb}; and $\bm{c}_s$, $\bm{c}_f$, $\bm{c}_r$ are the context vectors produced through the token, frame, and role channels, respectively.

\begin{figure*}[!t]
\begin{center}
\includegraphics[width=120mm]{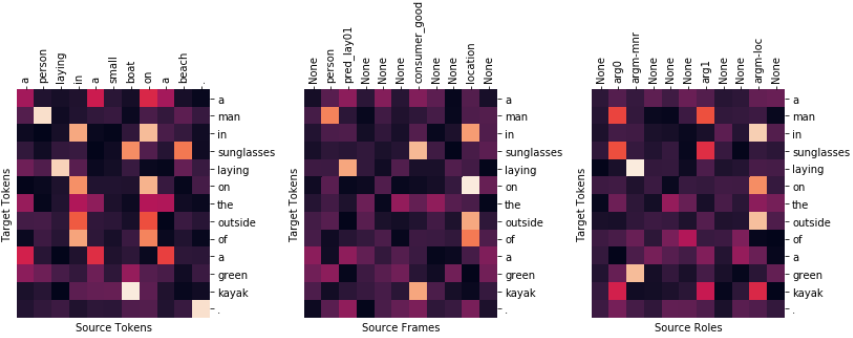}
\caption{{\small Example for target$\rightarrow$source attention (tokens, frames and roles). Each role is an attention weight distribution over source tokens/frames/roles (normalized). Attention weights taken from the first layer of the \textsc{transformer-pb}.}}\label{fig:attention}
\end{center}
\end{figure*}

\section{Automatic Evaluation}
\label{sec:auto-eval}

We first use easily computed automatic metrics to evaluate Transformers and semantic augmentation of both Transformers and \textsc{sr-lstm}. We train and assess the models on MSCOCO and WikiAnswers with the same setup as in previous work. In particular, for a fair and normalized comparison, we control factors such as computational machinery (including hardware and software), random seeding, hyperparameter tuning, etc. by implementing all the models in the same toolkits.%
\footnote{\ct{Crane:18} argues convincingly that factors such as hardware (e.g. GPU type and configuration), software (framework and version) and random seeding have nontrivial impacts on results.}%
\ Specifically, we build all models with PyTorch 0.3.1 (hyperparameters tuned to their best performance with the same random seeds) and run all experiments on a Titan Xp GPU. All models are tuned to their best performing configuration, and the average over 20 runs is recorded. For all models we apply greedy decoding.%
\footnote{Rather than beam search, which shows little performance variation from greedy search, as observed in our experiments and reported in the original works \cite{Prakash:16,Gupta:18}.}%
\ All word embeddings are initialized with 300D \texttt{GloVe} vectors \cite{Pennington:14}. For \textsc{sr-lstm}, \textsc{nv-lstm} and \textsc{transformer} respectively: the learning rates are 1e-4/1e-5/1e-5; the dropout rates are 0.5/0.3/0.1. For the \textsc{transformer}, we set the weight decay factor at 0.99, with the warmup at 500.

For the first experiment, following \ct{Prakash:16}, we restrict complexity by truncating all sentences to 15 words. We also gauge model speed as \emph{tokens per second}. As shown in Table \ref{tab:auto-results}, \textsc{transformer} performs comparably with the current state-of-the-art \textsc{nv-lstm}---and is nearly 7x faster with GPU. Also, while \textsc{sr-lstm} generally underperforms \textsc{nv-lstm}, our version enhanced with \texttt{SLING}-parsed semantic information (\textsc{sr-lstm-pb}) pulls ahead in all metrics except speed. This slowdown relative to other models is expected, since the multi-encoder increases model size. But \textsc{transformer-pb} parallelizes the added computations while retaining the relative performance gain, scoring the highest on all metrics. This experiment sets a new state-of-the-art for these benchmarks with about 6-7x speed-up.

\begin{table}[!t]
\begin{center}
\scalebox{0.85}{
\begin{tabular}{lcc}
Model & MSCOCO & WikiAnswers \\
\midrule
\textsc{sr-lstm} (ours) & 11.6 & 11.3 \\
\textsc{nv-lstm} (ours) & 10.9 & 12.1 \\
\textsc{transformer} & 17.7 & \textbf{16.8} \\
\textsc{sr-lstm-pb} (ours) & 11.5 & 11.6 \\
\textsc{transformer-pb} (ours) & \textbf{17.9} & 16.4 \\
\end{tabular}}
\caption{{\small \texttt{BLEU} scores on long sentences ($>$20 words). Similar patterns hold for \texttt{METEOR} and \texttt{TER}.}}\label{fig:long-sentences}
\end{center}
\end{table}

We also measured \texttt{BLEU} scores for models trained on all sentences but evaluated only on sentences with $>$20 words (Table \ref{fig:long-sentences}). The Transformer models prove especially effective for longer sentences, with a 4-6 \texttt{BLEU} margin over the LSTM models. Adding semantics did not provide further benefit on these metrics. This is likely in part because the quality of \texttt{SLING}'s SRL annotations goes down on longer sentences: \texttt{SLING}'s SPAN and ROLE F1 scores were respectively $1.5\%$ and $9.8\%$ lower on long sentences. Though \texttt{SLING}'s lower performance could be due partly to its in-built one-pass greedy decoder, long-range dependencies pose a general problem for all semantic role labeling systems. They employ mechanisms ranging from tweaking long sentences~\cite{Vickrey2008SentenceSF} to incorporating syntax~\cite[Sec.~A.1]{Strubell2018LinguisticallyInformedSF}. Exploring such approaches is a possible avenue for future work.

\begin{table}[!t]
\begin{center}
\scalebox{0.85}{
\begin{tabular}{lcccc}
\multirow{2}{*}{Model} & \multicolumn{4}{c}{MSCOCO Train Size} \\
\cmidrule{2-5}
& 50K & 100K & 200K & 331K (original) \\
\midrule
\textsc{sr-lstm} & 9.7 & 18.3 & 29.3 & 36.5 \\
\textsc{nv-lstm} & 5.3 & 15.4 & 25.5 & \textbf{41.1} \\
\textsc{transformer} & \textbf{9.9} & \textbf{24.5} & \textbf{33.9} & 41.0 \\
\end{tabular}}
\caption{{\small Experimenting with lower data sizes: performance (\texttt{BLEU}) of three models with randomly sampled training data (\texttt{MSCOCO}) in three sizes. Note that this evaluation is for all examples, with no length restrictions.}}\label{tab:training-size}
\end{center}
\end{table}

\begin{table}[!t]
\begin{center}
\scalebox{0.85}{
\begin{tabular}{lcccc}
Data & None & Frame-only & Role-only & Both \\
\midrule
MSCOCO & 41.0 & 42.5 & 41.8 & 44.0 \\
WikiAnswers & 41.9 & 43.2 & 43.0 & 43.9 \\
\end{tabular}}
\caption{{\small Ablation study (\texttt{BLEU}) with the best model \textsc{transformer-pb}. Note that this evaluation is for all examples, with no length restrictions.}}\label{tab:ablation}
\end{center}
\end{table}

As mentioned earlier, a key limitation of the seq2seq approach for paraphrasing is that it has orders of magnitude less data than typically available for training MT systems using the same model architecture. To see how training set size impacts performance, we trained and evaluated the three base systems on three randomly selected subsets of MSCOCO with 50K, 100K, and 200K examples each, plus using all 331K examples. Results in Table \ref{tab:training-size} show that successively doubling the number of examples gives large improvements in \texttt{BLEU}. The \textsc{transformer} model is particularly able to exploit additional data, attesting to its overall representational capacity given sufficient training examples. Note also that using all 331k examples in MSCOCO produces a \texttt{BLEU} score of 41.0---a strong indication that more data will likely further improve model quality.


\begin{table*}[t]
\begin{center}
\scalebox{0.85}{
\begin{tabular}{lll}
Type & Target & Prediction \\
\midrule
\multirow{2}{*}{Low-\texttt{BLEU}/Good} & a man in sunglasses laying on a green kayak. & the man laying on a boat in the water. \\
& a girl in a jacket and boots with a black umbrella. & a little girl holding a umbrella. \\
\cmidrule{2-3}
\multirow{2}{*}{Low-\texttt{BLEU}/Bad} & people on a gold course enjoy a few games & a group of people walking. \\
& a tall building and people walking & a large building with a clock on the top \\
\cmidrule{2-3}
\multirow{2}{*}{High-\texttt{BLEU}} & a picture of someone taking a picture of herself. & a woman taking a picture with a cell phone. \\
& a batter swinging a bat at a baseball. & a baseball player swinging a bat at a ball. \\
\end{tabular}}
\caption{{\small Examples from sample study. Pairs with Low-\texttt{BLEU} score $<$10 in \texttt{BLEU}, and those with High-\texttt{BLEU} score $>$25. In general, high \texttt{BLEU} correlates with good paraphrasing, but low \texttt{BLEU} includes both relatively good and bad paraphrasing.}}\label{tab:bad-bleu}
\end{center}
\end{table*}

Table \ref{tab:ablation} shows the results of ablating semantic information in the \textsc{transformer} model, with four conditions: None (i.e. \textsc{transformer}), Frame-only, Role-only and Both (i.e. \textsc{transformer-pb}). Frame and role information each improve results by 1-2 points on their own, but combining them yields the largest overall gain for both datasets.


We can also examine the \textsc{transformer-pb}'s attention alignments for indirect evidence of the contribution of semantics. 
Figure \ref{fig:attention} shows a cherry-picked example of target-to-source attention weights in the first layer of the multi-encoder.\footnote{We did not observe clear patterns for other layers.} For token-token attention, words are aligned, more or less as expected, with distributional lexical similarity \cite[for instance]{Luong:16}. For token-to-frame and token-to-role alignment, word tokens attend to frames and roles with interpretable relations. For example, \emph{man} has heavy attention weights for the frame \texttt{person} and the roles \texttt{arg0} and \texttt{arg1}; \emph{sunglasses} for the frame \texttt{consumer\_good}; \emph{laying} for \texttt{argm-mnr} (manner).

\section{Human Evaluation}
\label{sec:human-eval}

\begin{table*}[!t]
\begin{center}
\scalebox{0.85}{
\begin{tabular}{lll}
Target & Low-\texttt{BLEU} paraphrases & High-\texttt{BLEU} paraphrases \\
\midrule
several surfers are heading out into the waves. & some guys are running towards the ocean. & some surfers heading into the waves. \\
& [\texttt{BLEU} = 6.0] & [\texttt{BLEU} = 34.2] \\
a kitchen with wooden cabinets and tile flooring & 
wood closet in a kitchen with tile on floor & a kitchen with cabinets and flooring \\
& [\texttt{BLEU} = 17.7] & [\texttt{BLEU} = 27.0] \\
\end{tabular}}
\caption{{\small Examples illustrating disconnect between standard \texttt{BLEU} metric and human intuition: low-scoring paraphrases are more diverse than high-scoring ones, which tend toward parroting.}}\label{tab:parrot}
\end{center}
\end{table*}

\begin{table}[!t]
\begin{center}
\scalebox{0.85}{
\begin{tabular}{lcc}
& {\small {Task 1}} & {\small {Task 2}} \\
Model & {\small (target,pred)} & {\small (target,src,pred)} \\
\midrule
\textsc{nv-lstm} & 19.0 & 36.4 \\
\textsc{transformer} & 31.1 & 45.6 \\
\textsc{transformer-pb} & 36.4 & 66.5
\end{tabular}}
\caption{{\small Human acceptability scores for models presented with (target, prediction); and (target, source, prediction).}}\label{tab:coco}
\end{center}
\end{table}


\begin{table}[!t]
\begin{center}
\scalebox{0.85}{
\begin{tabular}{cc|c}
\textsc{transformer} & \textsc{transformer-pb} & \textsc{CHIA} \\
\hline
18.0 & 28.0 & 78.2
\end{tabular}}
\caption{{\small Similar-domain comparison on 1000 sampled \texttt{CHIA} images by 3 human evaluators, paired with predictions from the Transformer with and without semantic augmentation and with the gold CHIA captions.}}\label{tab:chia}
\end{center}
\end{table}

While adding semantic information significantly boosts performance on automated evaluation metrics, it is unclear whether the gains actually translate into higher-quality paraphrases. As noted above, MT metrics like \texttt{BLEU} reward fidelity to the input phrase, unlike human judgments that value lexical and syntactic diversity. This systematic divergence means that reasonable paraphrases can receive low \texttt{BLEU} scores (Table \ref{tab:bad-bleu}); and conversely that high \texttt{BLEU} scores can be obtained by parroting of the input sentence (Table \ref{tab:parrot}).

A related question is: how can we be sure our models can generalize to other datasets and domains? MSCOCO's caption-based data grants considerable semantic license since the ``paraphrase'' caption can include more or less or different detail than the original. Paraphrases involve several kinds of variation, and overall paraphrase acceptability is a matter of degree. Are our models learning robust generalizations about what kinds of variation preserve meaning? 

\begin{table*}
\scalebox{0.9}{
\begin{tabular}{l}
1-O: {The figure is illuminated by four footlights in the base and its proper left arm is raised.}\\
1-T: {Baseball players in a field playing baseball on a field.}\\
\midrule
2-O: {One day, while going to bed, Tara tells Ramesh that the paw works as she'd wished to solve the problem.}\\
2-T: {Photograph of a man in a suit and tie with a hat on his head.}\\
\midrule
3-O: {According to free people, democracy will not work without the element of the sensitive, conscience of an individual.}\\
3-T: {Photograph of a man in a suit and tie looking at a cell phone.}\\
\end{tabular}}
\caption{{\small Paraphrase predictions on out-of-domain (Wikipedia) sentences (O=Original, T=Transformer), demonstrating (1) hallucination of a baseball scene; and (2 and 3) bias toward predicting \emph{photograph of a man in a suit and tie} for novel input. Note that n-grams in the bias sequence appear frequently in the corpus (e.g. \emph{in a suit and tie} ranks 345 out of over 21M unique 5-grams), accounting for the failure mode of ``backing off'' to high-frequency sequences when testing on out-of-domain input.}}\label{tab:suit-tie}
\end{table*}

To investigate these issues, we turned to crowdsourced human judgments. We ran several experiments that varied in dataset used (same domain, similar domain, out-of-domain) and paraphrase decision (roughly, what kinds of inputs are presented). All cases require a binary judgment, averaged over multiple human evaluators. Each is described below.

\medskip \noindent \textbf{Task 1. Same domain, standard paraphrase:} Decide whether $s1$ and $s2$ have approximately the same meaning.

This task compares \textsc{nv-lstm} (the previous state-of-the-art) with our \textsc{transformer} and \textsc{transformer-pb} models on the standard paraphrase task. We randomly sampled 100 sentences from MSCOCO, paired them with their predicted paraphrases from each model, and presented each to 5 human evaluators. Results in Table~\ref{tab:coco}(a) demonstrate that our models significantly outperformed the state-of-the-art, with the \textsc{transformer-pb} hitting 36.4\% paraphrase acceptability, compared to \textsc{nv-lstm}'s 19.0\%.

Given the relatively low bar, however, it is worth considering how well the task as presented fits the data. As noted earlier, MSCOCO ``paraphrases'' are based on multiple descriptions of the same image; it thus includes many pairs that differ significantly in details or focus. For example, \textit{a boy standing on a sidewalk by a skateboard} and \textit{there is a young boy that is standing between some trees} differ in details to raise doubt about whether they mean ``the same thing''---but they are nonetheless compatible descriptions of the same scene. A judgment better matched to the annotated data could provide more insight into model performance.

\medskip \noindent \textbf{Task 2. Same domain, pair-based paraphrase:} Assuming $s1$ and $s2$ have approximately the same meaning, decide whether $s3$ also has approximately the same meaning.

This task adapts the standard paraphrase judgment to better handle the inherent variability of caption-based training data. Rather than explicitly define what is meant by ``approximately the same meaning'', we give raters additional context by providing both source and target (i.e. a gold training pair) as input to the task. This approach flexibly accommodates a wide range of differences between source and target, but (implicitly) requires them to be mutually consistent.

As shown in Table~\ref{tab:coco}(b), all systems score much higher for this evaluation over standard paraphrase task. Interestingly, we see even greater gains of \textsc{transformer-pb} (66.5\%) over \textsc{transformer} (45.6\%) for this task.

\medskip \noindent \textbf{Task 3. Similar domain, image-based paraphrase:} Given image $i$ and its caption $s1$, decide whether a paraphrase $s2$ generated from $s1$ is an acceptable caption for $i$.

To address the key issue of domain generalization, we evaluated model performance on a very near neighbor: CHIA, a collection of 3.3M image caption pairs of a specific caption with a more general (paraphrase) caption \cite{Sharma:18}.\footnote{CHIA captions are automatically modified versions of the original alt-text associated with the images, e.g. \textit{Musician Justin Timberlake performs at the 2017 Pilgrimage Music \& Cultural Festival on September 23, 2017 in Franklin, Tennessee} is generalized to \textit{pop artist performs at the festival in a city} as the CHIA caption.}\ We sampled 1,000 CHIA images and presented image-caption pairs to 3 human raters, where captions come from our \textsc{transformer}, \textsc{transformer-pb} models and the gold \textsc{CHIA} paraphrase.


Table~\ref{tab:chia} shows a large gain for the semantically enhanced \textsc{transformer-pb} (28.0\%) over the basic \textsc{transformer} (18.0\%)---but both fall far short of CHIA (78.2\%).

\medskip
\textbf{Out-of-domain, standard paraphrase.} The situation grows more dire as we venture further out of domain to tackle Wikipedia data. Our models fail so miserably that we did not bother with human evaluation, with most predictions hallucinatory at best (Table \ref{tab:suit-tie}). A striking behavior of the basic \textsc{Transformer} model was that it frequently resorts to mentioning men in suits and ties when presented with novel inputs (sentences 2 and 3).

There are some qualitative indications that the \textsc{transformer-pb} makes somewhat more thematically relevant predictions. For example:

\begin{itemize}

\item \textsc{Original}: \emph{In this fire, he lost many of his old notes and books including a series on Indian birds that he had since the age of 3}
\item \textsc{transformer}: \emph{bird sitting on a table with a laptop computer on it 's side of the road and a sign on the wall}
\item \textsc{transformer-pb}: \emph{yellow fire hydrant near a box of birds on a table with books on it nearby table}. 
\end{itemize}

\noindent
However, neither model's attempt is even close to adequate. 


\medskip 
Taken as a whole, human evaluations (both quantitative and qualitative) provide a more nuanced picture of how our models perform. Across all tasks, we find consistent improvements for models that include semantic information. But none of the models demonstrates much ability to generalize to out-of-domain examples.


\section{Conclusion}
\label{sec:conclusion}

We have shown substantial improvements in paraphrase generation through a simple semantic augmentation strategy that works equally well with both LSTMs and Transformers. Much scope for future work remains. Since \texttt{SLING} is a neural model, it associates both semantic frames and roles with continuous vectors, and the parser state upon adding frames/roles to the graph can serve as a continuous internal representation of the input's incrementally constructed meaning. In future work we plan to integrate this continuous representation with the symbolic (final) frame graph used here, as well as train \texttt{SLING} for other tasks.

Our evaluations also suggest a fundamental mismatch between automated metrics borrowed from MT, which are optimized for surface similarity, and the more nuanced factors that affect human judgment of paraphrase quality. Automated metrics, while far more scalable in speed and cost, cannot (yet) substitute for human evaluation. Paraphrase generation research would benefit from better use of existing datasets \cite{Iyer:17,xu-callisonburch-dolan:2015:SemEval,lan2017continuously}, creation of new datasets, and evaluation of metrics similar to that done by \ct{toutanova-EtAl:2016:EMNLP2016} for abstractive compression.

In particular, crowdsourcing tasks that provide more evidence about what is meant by ``meaning''---such as additional input instances, or the source image---may better accommodate diverse sources of paraphrase data. Future work can explore how robustly models perform given different amounts and types of data from a variety of tasks.

{\small
\bibliography{aaai}}

\begin{thebibliography}{}

\bibitem[\protect\citeauthoryear{Ba, Kiros, and Hinton}{2016}]{Ba:16}
Ba, J.~L.; Kiros, J.~R.; and Hinton, G.~E.
\newblock 2016.
\newblock {Layer Normalization}.
\newblock In {\em CoRR}.

\bibitem[\protect\citeauthoryear{Bahdanau, Cho, and Bengio}{2015}]{Bahdanau:15}
Bahdanau, D.; Cho, K.; and Bengio, Y.
\newblock 2015.
\newblock {Neural Machine Translation by Jointly Learning to Align and
  Translate}.
\newblock In {\em Proceedings of ICLR}.

\bibitem[\protect\citeauthoryear{Bannard and Callison-Burch}{2005}]{Bannard:05}
Bannard, C., and Callison-Burch, C.
\newblock 2005.
\newblock {Paraphrasing with bilingual parallel corpora}.
\newblock In {\em Proceedings of ACL}.

\bibitem[\protect\citeauthoryear{Barzilay and McKeown}{2001}]{Barzilay:01}
Barzilay, R., and McKeown, K.
\newblock 2001.
\newblock {Extracting paraphrases from a parallel corpus}.
\newblock In {\em Proceedings of ACL}.

\bibitem[\protect\citeauthoryear{Berant and
  Liang}{2014}]{berant-liang:2014:P14-1}
Berant, J., and Liang, P.
\newblock 2014.
\newblock Semantic parsing via paraphrasing.

\bibitem[\protect\citeauthoryear{Bowman \bgroup et al\mbox.\egroup
  }{2015}]{Bowman:15}
Bowman, S.~R.; Vilnis, L.; Vinyals, O.; Dai, A.~M.; Jozefowicz, R.; and Bengio,
  S.
\newblock 2015.
\newblock {Generating Sentences from a Continuous Space}.
\newblock In {\em CoRR}.

\bibitem[\protect\citeauthoryear{Callison-Burch, Cohn, and
  Lapata}{2008}]{Callison-Burch:08}
Callison-Burch, C.; Cohn, T.; and Lapata, M.
\newblock 2008.
\newblock {Parametric: An automatic evaluation metric for paraphrasing}.
\newblock In {\em Proceedings of COLING}.

\bibitem[\protect\citeauthoryear{Chaganty, Mussmann, and
  Liang}{2018}]{Chaganty:18}
Chaganty, A.~T.; Mussmann, S.; and Liang, P.
\newblock 2018.
\newblock {The Price of Debiasing Automatic Metrics in Natural Language
  Processing}.
\newblock In {\em Proceedings of ACL}.

\bibitem[\protect\citeauthoryear{Chen and Dolan}{2011}]{Chen:11}
Chen, D.~L., and Dolan, W.~B.
\newblock 2011.
\newblock {Collecting Highly Parallel Data for Paraphrase Evaluation}.
\newblock In {\em Proceedings of ACL}.

\bibitem[\protect\citeauthoryear{Crane}{2018}]{Crane:18}
Crane, M.
\newblock 2018.
\newblock {Questionable Answers in Question Answering Research: Reproducibility
  and Variability of Published Results}.
\newblock In {\em Proceedings of TACL}.

\bibitem[\protect\citeauthoryear{Ellsworth and Janin}{2007}]{Ellsworth:07}
Ellsworth, M., and Janin, A.
\newblock 2007.
\newblock {Mutaphrase: Paraphrase with FrameNet}.
\newblock In {\em Proceedings of ACL}.

\bibitem[\protect\citeauthoryear{Fader, Zettlemoyer, and
  Etzioni}{2013}]{Fader:13}
Fader, A.; Zettlemoyer, L.; and Etzioni, O.
\newblock 2013.
\newblock {Paraphrase-Driven Learning for Open Question Answering}.
\newblock In {\em Proceedings of ACL}.

\bibitem[\protect\citeauthoryear{Fader, Zettlemoyer, and
  Etzioni}{2014}]{Fader:14}
Fader, A.; Zettlemoyer, L.; and Etzioni, O.
\newblock 2014.
\newblock {Open Question Answering Over Curated and Extracted Knowledge Bases}.
\newblock In {\em Proceedings of KDD}.

\bibitem[\protect\citeauthoryear{Fillmore}{1982}]{fillmore82:_frame}
Fillmore, C.~J.
\newblock 1982.
\newblock {\em Frame semantics}.
\newblock Seoul, South Korea: Hanshin Publishing Co.
\newblock  111--137.

\bibitem[\protect\citeauthoryear{Gupta \bgroup et al\mbox.\egroup
  }{2018}]{Gupta:18}
Gupta, A.; Agarwal, A.; Singh, P.; and Rai, P.
\newblock 2018.
\newblock {A Deep Generative Framework for Paraphrase Generation}.
\newblock In {\em Proceedings of AAAI}.

\bibitem[\protect\citeauthoryear{He \bgroup et al\mbox.\egroup }{2015}]{He:15}
He, K.; Zhang, X.; Ren, S.; and Sun, J.
\newblock 2015.
\newblock { Deep Residual Learning for Image Recognition}.
\newblock In {\em CoRR}.

\bibitem[\protect\citeauthoryear{He, Lewis, and Zettlemoyer}{2015}]{He:2015}
He, L.; Lewis, M.; and Zettlemoyer, L.~S.
\newblock 2015.
\newblock Question-answer driven semantic role labeling: Using natural language
  to annotate natural language.
\newblock In {\em EMNLP}.

\bibitem[\protect\citeauthoryear{Issa \bgroup et al\mbox.\egroup
  }{2018}]{issa:etal:2018}
Issa, F.; Damonte, M.; Cohen, S.~B.; Yan, X.; and Chang, Y.
\newblock 2018.
\newblock Abstract meaning representation for paraphrase detection.
\newblock In {\em Proceedings of NAACL-HLT18},  486--492.

\bibitem[\protect\citeauthoryear{Iyer, Dandekar, and Csernai}{2017}]{Iyer:17}
Iyer, S.; Dandekar, N.; and Csernai, K.
\newblock 2017.
\newblock {First Quora Dataset Release: Question Pairs}.
\newblock In {\em \url{data.quora.com}}.

\bibitem[\protect\citeauthoryear{Kingma and Welling}{2014}]{Kingma:14}
Kingma, D.~P., and Welling, M.
\newblock 2014.
\newblock {Auto-encoding Variational Bayes}.
\newblock In {\em Proceedings of ICLR}.

\bibitem[\protect\citeauthoryear{Kozlowski, McCoy, and
  Vijay-Shanker}{2003}]{Kozlowski:03}
Kozlowski, R.; McCoy, K.~F.; and Vijay-Shanker, K.
\newblock 2003.
\newblock {Generation of Single-sentence Paraphrases from Predicate/Argument
  Structure Using Lexico-Grammatical Resources}.
\newblock In {\em Proceedings of IWP}.

\bibitem[\protect\citeauthoryear{Lan \bgroup et al\mbox.\egroup
  }{2017}]{lan2017continuously}
Lan, W.; Qiu, S.; He, H.; and Xu, W.
\newblock 2017.
\newblock A continuously growing dataset of sentential paraphrases.
\newblock In {\em Proceedings of The 2017 Conference on Empirical Methods on
  Natural Language Processing (EMNLP)}.

\bibitem[\protect\citeauthoryear{Lavie and Agarwal}{2007}]{Lavie:07}
Lavie, A., and Agarwal, A.
\newblock 2007.
\newblock {Meteor: An Automatic Metric for MT Evaluation with High Levels of
  Correlation with Human Judgments}.
\newblock In {\em Proceedings of SMT}.

\bibitem[\protect\citeauthoryear{Lin \bgroup et al\mbox.\egroup
  }{2014}]{Lin:14}
Lin, T.-Y.; Maire, M.; Belongie, S.; Bourdev, L.; Girshick, R.; Hays, J.;
  Perona, P.; Ramanan, D.; Zitnick, C.~L.; and Doll\'ar, P.
\newblock 2014.
\newblock {Microsoft COCO: Common Objects in Context}.
\newblock In {\em CoRR}.

\bibitem[\protect\citeauthoryear{Liu, Dahlmeier, and Ng}{2010}]{Liu:10}
Liu, C.; Dahlmeier, D.; and Ng, H.~T.
\newblock 2010.
\newblock {PEM: A paraphrase evaluation metric exploiting parallel texts}.
\newblock In {\em Proceedings of EMNLP}.

\bibitem[\protect\citeauthoryear{Luong, Pham, and Manning}{2016}]{Luong:16}
Luong, M.-T.; Pham, H.; and Manning, C.~D.
\newblock 2016.
\newblock {Effective Approaches to Attention-based Neural Machine Translation}.
\newblock In {\em Proceedings of EMNLP}.

\bibitem[\protect\citeauthoryear{Madnani and Dorr}{2010}]{Madnani:10b}
Madnani, N., and Dorr, B.~J.
\newblock 2010.
\newblock {Generating Phrasal and Sentential Paraphrases: A Survey of
  Data-driven Methods}.
\newblock In {\em Computational Linguistics, 36(3)}.

\bibitem[\protect\citeauthoryear{Madnani and Tetreault}{2010}]{Madnani:10a}
Madnani, N., and Tetreault, J.
\newblock 2010.
\newblock {Re-examining Machine Translation Metrics for Paraphrase
  Identification}.
\newblock In {\em Proceedings of NAACL}.

\bibitem[\protect\citeauthoryear{Marcheggiani, Bastings, and
  Titov}{2018}]{marcheggiani:etal:2018}
Marcheggiani, D.; Bastings, J.; and Titov, I.
\newblock 2018.
\newblock Exploiting semantics in neural machine translation with graph
  convolutional networks.
\newblock In {\em Proceedings of NAACL-HLT18},  486--492.

\bibitem[\protect\citeauthoryear{McKeown}{1983}]{McKeown:83}
McKeown, K.~R.
\newblock 1983.
\newblock {Paraphrasing Questions Using Given and New Information}.
\newblock In {\em American Journal of Computational Linguistics}.

\bibitem[\protect\citeauthoryear{Papineni \bgroup et al\mbox.\egroup
  }{2002}]{Papineni:02}
Papineni, K.; Roukos, S.; Ward, T.; and Zhu, W.-J.
\newblock 2002.
\newblock {BLEU: a Method for Automatic Evaluation of Machine Translation}.
\newblock In {\em Proceedings of ACL}.

\bibitem[\protect\citeauthoryear{Pavlick \bgroup et al\mbox.\egroup
  }{2015}]{pavlick:etal:2015}
Pavlick, E.; Bos, J.; Nissim, M.; Beller, C.; Van~Durme, B.; and
  Callison-Burch, C.
\newblock 2015.
\newblock Adding semantics to data-driven paraphrasing.

\bibitem[\protect\citeauthoryear{Pennington, Socher, and
  Manning}{2014}]{Pennington:14}
Pennington, J.; Socher, R.; and Manning, C.~D.
\newblock 2014.
\newblock {GloVe: Global Vectors for Word Representation}.
\newblock In {\em Proceedings of EMNLP}.

\bibitem[\protect\citeauthoryear{Prakash \bgroup et al\mbox.\egroup
  }{2016}]{Prakash:16}
Prakash, A.; ; Hasan, S.~A.; Lee, K.; Datla, V.; Qadir, A.; Liu, J.; and Farri,
  O.
\newblock 2016.
\newblock {Neural Paraphrase Generation with Stacked Residual LSTM Networks}.
\newblock In {\em Proceedings of COLING}.

\bibitem[\protect\citeauthoryear{Ringgaard, Gupta, and
  Pereira}{2017}]{Ringgaard:17}
Ringgaard, M.; Gupta, R.; and Pereira, F. C.~N.
\newblock 2017.
\newblock {SLING: A Framework for Frame Semantic Parsing}.
\newblock In {\em CoRR}.

\bibitem[\protect\citeauthoryear{Schneider and Wooters}{2018}]{Schneider:18}
Schneider, N., and Wooters, C.
\newblock 2018.
\newblock {The NLTK FrameNet API}.
\newblock In {\em Proceedings of NAACL}.

\bibitem[\protect\citeauthoryear{Shah \bgroup et al\mbox.\egroup
  }{2018}]{Shah:18}
Shah, P.; Hakkani-Tur, D.; T\"ur, G.; Rastogi, A.; Bapna, A.; Nayak, N.; and
  Heck, L.
\newblock 2018.
\newblock {Building a Conversational Agent Overnight with Dialogue Self-Play}.
\newblock In {\em CoRR}.

\bibitem[\protect\citeauthoryear{Sharma \bgroup et al\mbox.\egroup
  }{2018}]{Sharma:18}
Sharma, P.; Ding, N.; Goodman, S.; and Soricut, R.
\newblock 2018.
\newblock {Conceptual Captions: A Cleaned, Hypernymed, Image Alt-text Dataset
  For Automatic Image Captioning}.
\newblock In {\em Proceedings of ACL}.

\bibitem[\protect\citeauthoryear{Snover \bgroup et al\mbox.\egroup
  }{2006}]{Snover:06}
Snover, M.; Dorr, B.; Schwartz, R.; Micciulla, L.; and Makhoul, J.
\newblock 2006.
\newblock {A Study of Translation Edit Rate with Targeted Human Annotation}.
\newblock In {\em Proceedings of AMT}.

\bibitem[\protect\citeauthoryear{Strubell \bgroup et al\mbox.\egroup
  }{2018}]{Strubell2018LinguisticallyInformedSF}
Strubell, E.; Verga, P.; Andor, D.; Weiss, D.; and McCallum, A.
\newblock 2018.
\newblock Linguistically-informed self-attention for semantic role labeling.
\newblock In {\em Proceedings of EMNLP}.

\bibitem[\protect\citeauthoryear{Toutanova \bgroup et al\mbox.\egroup
  }{2016}]{toutanova-EtAl:2016:EMNLP2016}
Toutanova, K.; Brockett, C.; Tran, K.~M.; and Amershi, S.
\newblock 2016.
\newblock A dataset and evaluation metrics for abstractive compression of
  sentences and short paragraphs.
\newblock In {\em Proceedings of EMNLP}.

\bibitem[\protect\citeauthoryear{Vaswani \bgroup et al\mbox.\egroup
  }{2017}]{Vaswani:17}
Vaswani, A.; Shazeer, N.; Parmar, N.; Uszkoreit, J.; Jones, L.; Gomez, A.~N.;
  Kaiser, L.; and Polosukhin, I.
\newblock 2017.
\newblock {Attention is All You Need}.
\newblock In {\em Proceedings of NIPS}.

\bibitem[\protect\citeauthoryear{Vickrey and
  Koller}{2008}]{Vickrey2008SentenceSF}
Vickrey, D., and Koller, D.
\newblock 2008.
\newblock Sentence simplification for semantic role labeling.
\newblock In {\em Proceedings of ACL}.

\bibitem[\protect\citeauthoryear{Wubben, van~den Bosch, and
  Krahmer}{2010}]{Wubben:10}
Wubben, S.; van~den Bosch, A.; and Krahmer, E.
\newblock 2010.
\newblock {Paraphrase Generation as Monolingual Translation: Data and
  Evaluation}.
\newblock In {\em Proceedings of INLG}.

\bibitem[\protect\citeauthoryear{Xiong \bgroup et al\mbox.\egroup
  }{2018}]{Xiong:18}
Xiong, H.; He, Z.; Hu, X.; and Wu, H.
\newblock 2018.
\newblock {Multi-channel Encoder for Neural Machine Translation}.
\newblock In {\em Proceedings of AAAI}.

\bibitem[\protect\citeauthoryear{Xu, Callison-Burch, and
  Dolan}{}]{xu-callisonburch-dolan:2015:SemEval}
Xu, W.; Callison-Burch, C.; and Dolan, B.
\newblock Semeval-2015 task 1: Paraphrase and semantic similarity in twitter
  (pit).
\newblock In {\em Proceedings of SemEval 2015}.

\bibitem[\protect\citeauthoryear{Zhao \bgroup et al\mbox.\egroup
  }{2009}]{Zhao:09}
Zhao, S.; Lan, X.; Liu, T.; and Li, S.
\newblock 2009.
\newblock {Application-driven Statistical Paraphrase Generation}.
\newblock In {\em Proceedings of ACL-IJCNLP}.

\end{thebibliography}
\bibliographystyle{aaai}

\end{document}